 \pdfoutput=1 %for arxiv
\documentclass{llncs}
\usepackage{graphicx}
\usepackage[utf8]{inputenc}
\usepackage[usenames]{color}
\usepackage{wrapfig} %
\usepackage{placeins} %
\usepackage{multirow}
\usepackage{url}

\def\ccalc{{\sc CCalc}}
\def\cplus{${\cal C}+$}

\def\ba{\begin{array}}
\def\ea{\end{array}}

\newcommand{\causes}[2]{\ensuremath{{#1\ {\textrm{\bf causes\ }}#2 }}}
\newcommand{\nonexecutableIf}[2]{\ensuremath{{\textrm{\bf nonexecutable\ }}#1  {\textrm{\bf\ if\ }} #2}}
\newcommand{\static}[2]{\ensuremath{{{\textrm{\bf caused\ }}#1{\textrm{\bf\ if\ }}#2}}}

\newcommand{\redtab}{\hspace{-0.8cm}}
\date{June 2013}

\begin{document}

    \title{Integration of 3D Object Recognition and Planning for Robotic Manipulation:\\A Preliminary Report}

  \author
         {Damien Jade Duff \and Esra Erdem \and Volkan Patoglu\\
}

\institute{Sabanc\i\ University, \.Istanbul, Turkey\\
\email{\{damienjduff,esraerdem,vpatoglu\}@sabanciuniv.net}
}
\label{firstpage}

\maketitle

  \begin{abstract}
We investigate different approaches to integrating object
recognition and planning in a tabletop manipulation domain with the
set of objects used in the 2012 RoboCup@Work competition. Results of
our preliminary experiments show that, with some approaches, close
integration of perception and planning improves the quality of plans, as well
as the computation times of feasible plans.

  \end{abstract}


\section{Introduction}

Consider what the eyes are doing when involved in the solving of a
jigsaw puzzle. While the mind is darting about, imagining placement
possibilities, considering combinations, and pondering strategies,
the eyes too are darting from place to place over the puzzle,
examining pieces relevant to a considered placement, checking edges
for compatibility, and studying the layout. The eyes are responding
to the deliberation of the mind, checking expectations and seeking
out necessary information. They are assuring that the deliberation
is rooted in the physical reality of the problem.

Many problems in robot manipulation require deliberation because of
the large number of elements involved and their complicated
interactions. It
should be expected that in such problems a similar integration of
deliberation with perception (and geometric reasoning) would be
measurably beneficial, considering its apparent interleaving in
human problem solving.  Consider, for instance, moving objects around from
one location to another on a workspace. These movements must take into account
(1) whether some objects block the reachability of another object %
at a particular location, and (2) whether an object can be
placed on top of another object while maintaining the stability of the
stack of objects. To check the two conditions above, perception can
be useful in identifying the orientation and shape of the object; in
this way, perception can guide planning towards feasible plans.

Alternatively, planning may guide perception: rather than obtaining
the details of all the objects in the scene, their shapes and so
forth, planning may ask for the information about the relevant
objects thus reducing the amount of computation for perception,
and may further reduce the amount of perceptual knowledge needed to be considered by the planning.
Such a top-down guidance of perception can be considered a rudimentary attentional
mechanism. %

In this paper, we investigate the usefulness of three different
approaches to integrating perception with planning, in a similar way to Schüller~et~al.~2013~\cite{schuller_systematic_2013} who investigate integration of planning
with geometric reasoning.
The three approaches investigated are (\textsc{Pre}) preprocessing of perceptual data
and its integration into the action domain, (\textsc{Filt}) filtering of plans
by post-checking using perceptual processing, and (\textsc{Repl}) derivation of additional constraints from perception
for subsequent replanning.

\begin{figure}[b]
\centering
\includegraphics[angle=180,scale=0.05]{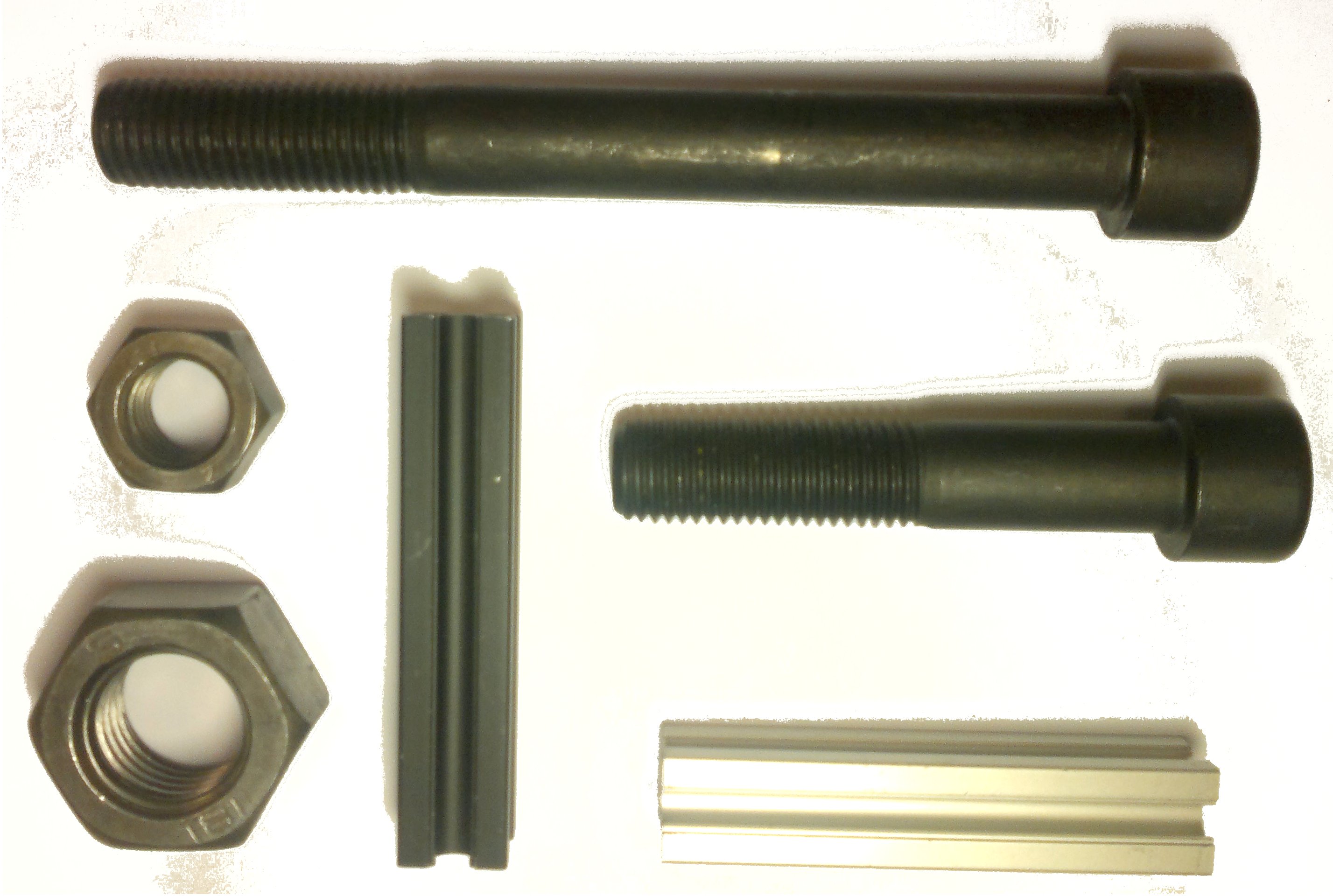}
 \caption{A subset of objects from the 2012 Robocup@Work mobile manipulation competition.}
 \label{atwork_objects}
\end{figure}

In the current work we describe experiments in a robotic
manipulation domain with various industrially plausible
objects used at the 2012 RoboCup@Work competition (Figure
\ref{atwork_objects}).

As with Erdem~et~al.~2011~\cite{erdem_combining_2011}, we describe the manipulation
domain in the action language
\cplus~\cite{giunchiglia_nonmonotonic_2004}, and use the reasoning
system \ccalc~\cite{mccain_causal_1997} to solve planning problems.
We use an object segmentation and shape recognition system built on
the Kinect RGBD (RGB plus depth) camera and Point Cloud Library~\cite{rusu_3d_2011}.
Perceptual processing does a quick initial
bottom-up run, finding candidate objects in the target scene, and
subsequently provides information about the shape of these objects to the planner
on demand.

\section{Related Work}

Various planning techniques have been
used  for efficient visual processing management in earlier studies~\cite{Gong_1992,Clouard_1999,Chien_2000,Thonnat_2000}. A survey of such works can be found in the context of the recent introduction of planning for perception in the context of cognitive robotics \cite{sridharan_planning_2010}.
The current report distinguishes itself in that it is an empirical investigation of embedding of perceptual processing into a task planning problem,
rather than an application of planning to perceptual processing and also aims to investigate how this integration might also improve efficiency of planning. In that sense,
a more relevant related work reports
a Prolog-based decision making system that utilizes external
computations for generating and evaluating perceptual hypotheses, such as the missing
objects on a table \cite{pangercic_combining_2010}, though it should be emphasized that the current report is an empirical investigation of different ways of embedding such external computations.

\section{Manipulation Domain Description}

The robotic manipulation domain we consider in our experiments
involves robot grasping, transport and placement of objects from a
small set of industrially plausible objects (Figure
\ref{atwork_objects}) used at the Robocup@Work 2012 competition. The
planning problem is to obtain a sequence of actions that transforms
an initial configuration of these objects on a work area into a
final configuration that satisfies some goal conditions. Perception
is utilized in identifying the shape of objects to check the
stability of stacks of objects and also their reachability.

In this work, to emphasize the integration of perception with
planning, unlike at Robocup@Work 2012, the mobile aspect of the
robot is not addressed.

We view the work area as a 5$\times$3 grid, where each grid cell has
a unique label. We assume that the objects are oriented in three
different ways: horizontal to the $x$ axis, horizontal to the $y$ axis,
or vertical with respect to the work area. Objects can be placed on top of each
other or on the work area.

We describe this domain in the action language
\cplus~\cite{giunchiglia_nonmonotonic_2004} as follows.

States of the world are described by means of two fluents (i.e.,
atoms whose truth value may change over time): one describing the
locations $loc$ of objects $obj$ on the work area or on other objects
($is\_at(obj)=loc$), and the other describing their orientations
$orient$ ($ori\_is(obj)=orient$).

We represent an action of the robot moving an object $obj$ to a
location $loc$ with an orientation $orient$ by atoms of the form
$move(obj,loc,orient)$. This action involves both a pick of the
object and its placement. In the following, $obj,obj'$ range over objects, $loc$ ranges over
locations on the work area (i.e., grid cells or other objects), and $orient,orient'$
ranges over the three possible orientations of objects.

\subsection{Effects and preconditions of actions} The direct effects of the move action are
described in \cplus\ by the following causal laws:
$$\ba l
\causes{move(obj,loc,orient)}{is\_at(obj)=loc} \\
\causes{move(obj,loc,orient)}{ori\_is(obj)=orient} \ea$$

The preconditions of this action are described by causal laws as
well. For instance, we can represent that an object cannot be moved
to a location if it is already there, by the causal law:
$$\nonexecutableIf{move(obj,loc,orient)}{is\_at(obj)=loc}$$

The precomputation (\textsc{Pre}) approach to integration also makes use of causal
laws as a way of integrating external computation. The following
causal law expresses that  an object cannot be moved on top of
another object if that would lead to an unstable stack.
$$\ba l \nonexecutableIf{move(obj,obj',orient)}{ori\_is(obj')=orient'} \\
\qquad (\textrm{where}\ unstackable\_ext(obj,orient,obj',orient')\
\textrm{holds}) \ea$$

Here, the stability of a stack of objects is checked ``externally''
by the ``external predicate''
$unstackable\_ext(obj,orient,obj',orient')$, which utilizes
perception to obtain the shape of the object (though object shape is not represented
at the high level)
and then checks the
stability of the stack with respect to some geometric constraints which depend on object
shape.
An external predicate is a predicate whose truth value is determined by running arbitrary
computation, the details of which are not represented in the high-level formalism from
which it is accessed. %
External predicates are similarly utilized by Erdem et al. 2011 \cite{erdem_combining_2011}
to embed geometric reasoning into
preconditions of actions.

Similarly, $reach\_blocked\_ext(obj,loc,orient,obj',loc',orient')$,  %
another external predicate, is used to
determine whether an object $obj$ above a particular table location $loc$ and
orientation $orient$ will block a reach to a second object $obj'$ above
a different table location $loc'$ and orientation $orient'$. This external
predicate is used to forbid certain actions. In one case, if the
object to be moved is currently blocked by another object, the move
action is forbidden. In another, an object cannot be placed on an
unreachable table location or on another object that is above an
unreachable table location.

\subsection{Constraints}
State constraints ensure that two objects cannot be at the same
location and an object cannot be below itself:
$$\ba l \static{false}{is\_at(obj)=loc \wedge is\_at(obj')=loc}
\qquad
(obj \neq obj') \\
\static{false}{is\_below(obj,obj)} \\
\ea$$

Here, $is\_below(obj,obj')$ is a derived predicate defined in terms
of $is\_at(obj)=loc$.

\subsection{Planning}\label{sec:planning}

Given the domain description partially explained above,
we can solve planning problems using the reasoning system
\ccalc~\cite{mccain_causal_1997} by means of ``queries'' like the
following:

\begin{verbatim}
:- query
  maxstep :: 0..3;
  % Initial State
  0: is_at(sco2)=loc_0x0, ori_is(sco2)=vert,
     is_at(obj1)=loc_2x1, ori_is(obj1)=vert;
  % Goal
  maxstep: is_below(loc_0x0,obj1).
\end{verbatim}

This query asks for a shortest plan whose length is at most 3, for a
planning problem with:

\begin{itemize}
\item {\bf Initial state:} the object \verb#sco2# is placed on the work place at
location \verb#loc_0x0# with a vertical orientation and the object
\verb#obj1# is placed at location \verb#loc_2x1# with a vertical
orientation, and
\item {\bf Goal:} a configuration of objects
such that \verb#obj1# is above location \verb#loc_0x0#.
\end{itemize}

\section{Object Recognition}

Perceptual data comes in the form of a point cloud from a Kinect
RGBD camera, which uses a structured infrared projector and camera to pick out dense depth
as well as RGB images.
The camera is mounted on the robot at an angle to gain a wide view
of the workspace. The perceptual subsystem consists of two phases,
the bottom-up and top-down phases.

In the first, bottom-up phase, candidate scene %
objects are segmented on the basis of disconnectedness in 3D space, and located with
respect to a tabletop workspace. Orientation is determined by object principle extents. This requires a mapping of the perceptual data into the world coordinate system,
using robot localization and calibration of sensor extrinsic parameters.
The names of task-relevant objects
are given in the task specification and simple data association is performed based on Euclidean distance. Other objects %
have automatically generated names that are forwarded on to the task
planner.

The second, top-down phase is the object recognition phase, where
the perceptual component provides the shapes of objects on request.
It uses a database of CAD-like models and associated
reference point clouds, and the shape is maintained as an identifier mapping into this database. Further orientation refinement is possible after recognition. %
The recognition phase proceeds by matching a simple global object descriptor, calculated from colors and principle axes, between available 3D reference point clouds and candidate object point clouds. Local shape descriptors are also robustly matched: Fast Point Feature Histograms \cite{rusu_fast_2009}. Both global and local features are used because the small size of target objects and the sensor's distance from them make local features unreliable. %

\section{Integration of Task Planning and Perception}

We consider three approaches to integrating perception and task
planning, in a similar way as geometric reasoning and task planning
are integrated by \cite{schuller_systematic_2013}.

\subsection{Precomputation}

\textsc{Pre}: In the precomputation approach, first all
possible external computations (i.e., stability and reachability checks)
involving perceptual processing (i.e., identifying shapes of all
objects) are completed, and then the results of these computations
are represented by means of external predicates used in the domain
description. After that, feasible plans are computed.

For instance, shape information and information about possible
occlusions returned from the perceptual system can be represented as
Prolog facts as follows:

\begin{verbatim}
  shape_is_ext(sco2,bolt_m20_100).
  is_in_front_ext(loc_0x0,loc_1x0).
\end{verbatim}

which then can be utilized in defining external predicates as
follows:

\begin{verbatim}
  unstackable_ext(OBJ1,OBJ1_ORIENTATION,OBJ2,_):-
      shape_is_ext(OBJ2,aluprofil_f20_100_gray),
      OBJ1_ORIENTATION\=vert,
      shape_is_ext(OBJ1,bolt_m20_100).

  reach_blocked_ext(OBJ1,LOC1,ORI1,_,LOC2,_):-
      is_in_front_ext(LOC1,LOC2),
      shape_is_ext(OBJ1,bolt_m20_100),
      ORI1=vert.
\end{verbatim}

The first rule states that a bolt can't be stacked horizontally on
top of an aluminium profile and the second that a vertical bolt will
block a reach to any object behind it.

Because object shape is constant
across the time domain, we do not here consider perception to
produce only initial state (initial locations and orientations),
but also can be used in causal laws that apply
at all time points. %

\subsection{Filtering}

\textsc{Filt}: In the filtering approach, a plan is
computed first using the domain description without the causal laws
that depend on external predicates, and then stability and
reachability checks are performed to identify the feasibility of the
plan; if the plan is not feasible a different plan is computed.
This three step procedure is executed in a loop until a feasible plan is
computed.
In the \textsc{Filt} condition,
external checks are not integrated into the domain
description as external predicates as is done in \textsc{Pre} but are
instead used after planning to evaluate the correctness of planner output;
i.e. they are not formally part of the domain description. %

External computations check plan feasibility as follows:
\begin{itemize}%
 \item It is determined whether the computed plan attempts a stack, or a reach where an object may be blocking another.
 This provides a list of queries of one of the two forms: ``Does putting the object $obj$ at orientation $orient$ on top of $obj'$ when it is at orientation $orient'$ make the objects unstable?'' and
``Does the object $obj$ at orientation $orient$ and table location $loc$ block a reach action to the object $obj'$ when it is at orientation $orient'$ and table location $loc'$?''. Although not utilized in the domain theory, these queries have the same form as the external predicates referred to by the \textsc{Pre} method: The first query has the form
$unstackable\_ext(obj,orient,obj',orient')$ and the second query has the form
$reach\_blocked\_ext(obj,loc,orient,obj',loc',orient')$.

\item For each of these queries it is inferred which perceptual information is
   necessary to to answer it (it is determined for which objects to calculate shapes). The perception
  module is then queried for this information if it is not already cached.

\item The truth of the queries
   are ascertained with the new data. If no relevant unstackable or unreachable query returns true, the plan is deemed
  feasible.
\end{itemize}

\subsection{Replanning}

\textsc{Repl}: The replanning approach, as in the filtering
approach, also follows the three step procedure in a loop. However,
in the last step, the planning problem is constrained by new
information obtained by perceptual processing and the planner
restarted with the new planning problem.

Consider, for instance, the planning problem presented in
Section~\ref{sec:planning}. After the computation of an infeasible
plan and identification of reasons for its infeasibility, the
planning problem can be modified as follows, ensuring that no more
infeasible plans with the same reasons for infeasibility are computed:

\begin{verbatim}
:- query
  maxstep :: 0..3;
  % Initial State
  0: is_at(sco2)=loc_0x0, ori_is(sco2)=vert,
     is_at(obj1)=loc_2x1, ori_is(obj1)=vert;
  % Constraints
  T=<maxstep-1 ->> (T: -move(sco2,obj1,PUT_ORI));
  T=<maxstep-1 ->> (T: ori_is(sco2)=vert
                    ->> -move(sco3,obj1,PUT_ORI));
  T=<maxstep-1 ->> (T: (is_below(loc_3x1,sco2);
                    ori_is(sco2)=vert)
                    ->> -move(OBJ,loc_3x2,BLOCKED_ORI));
  T=<maxstep-1 ->> (T: (is_below(loc_3x1,sco2),
                   ori_is(sco2)=vert, is_below(loc_3x2,OBJ))
                   ->> -move(OBJ,TARGET_LOC,PUT_ORI));
  % Goal
  maxstep: is_below(loc_0x0,obj1).
\end{verbatim}

The first constraint expresses that object \verb#sco2# cannot
be stacked on top of object \verb#obj1#. The second
constraint expresses that the converse stack is not possible if
object \verb#sco2# is vertically oriented (because the
perception module knows that object \verb#sco2# is a bolt). The third constraint
expresses that it is not possible to move an object over location
\verb#loc_3x2# if object \verb#sco2# is at location
\verb#loc_3x1# and is vertical. The fourth constraint states that if
object \verb#sco2# is at location \verb#loc_3x1# and is
vertical, then no object above table location \verb#loc_3x2# is
moveable.

\section{Experimental Evaluation}

In our experiments, we consider the three different approaches to
integrating planning and perception as described above, considering
three problem instances. RGBD images representing the
perceptual input into the system for each of the three instances,
and the planning problem as it is presented to the planner after the
bottom-up perceptual processing, are presented in Figure
\ref{all_instances}.

\paragraph{Instance 1:}
The initial configuration of objects in Instance 1, visible on the
left of Figure \ref{all_instances}, consists of a vertically
oriented bolt at the front on the right, in front of a large nut. A
second bolt is visible on the left. The aim of this scenario is to
test the planner in minor clutter to see if gain is obtained from
integrating perception and planning. The task is to move the nut to
the center of the workspace (move \verb#obj1# from \verb#loc_0x1# to
\verb#loc_2x1#). This requires the bolt to be moved first, which
ultimately requires its shape to be computed to calculate the
reachability of the nut. An elided example of an expected output
plan:

\begin{verbatim}
  0: move(sco1,loc_3x0,horiz_y).
  1: move(obj1,loc_2x1,horiz_y).
\end{verbatim}

\begin{figure}
\begin{center}
\begin{tabular}{c@{\redtab}c@{\redtab}c}
\includegraphics[scale=0.3]{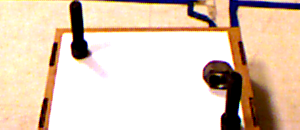}\redtab & \includegraphics[scale=0.3]{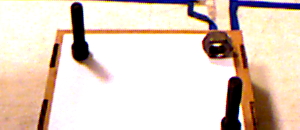}\redtab & \includegraphics[scale=0.3]{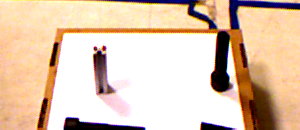} \\
\includegraphics[scale=0.3]{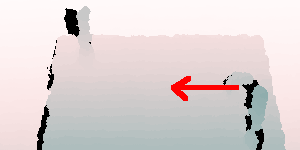}\redtab & \includegraphics[scale=0.3]{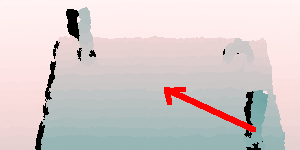}\redtab & \includegraphics[scale=0.3]{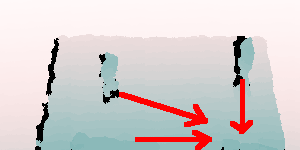} \\
\includegraphics[scale=0.6]{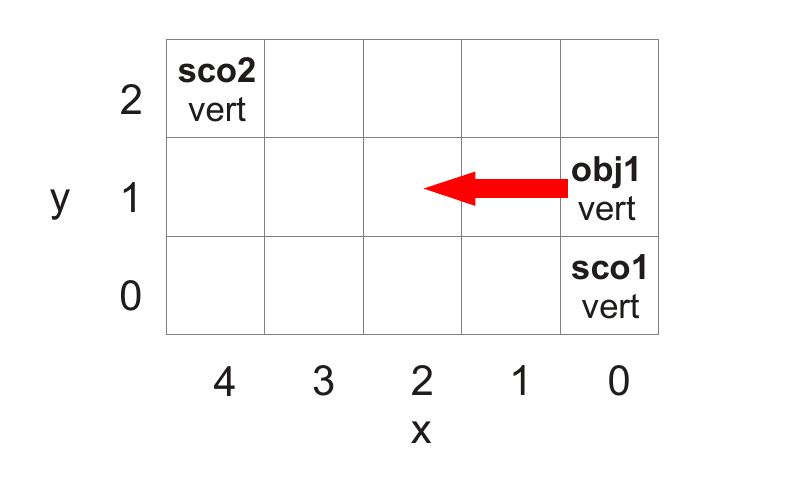}\redtab & \includegraphics[scale=0.6]{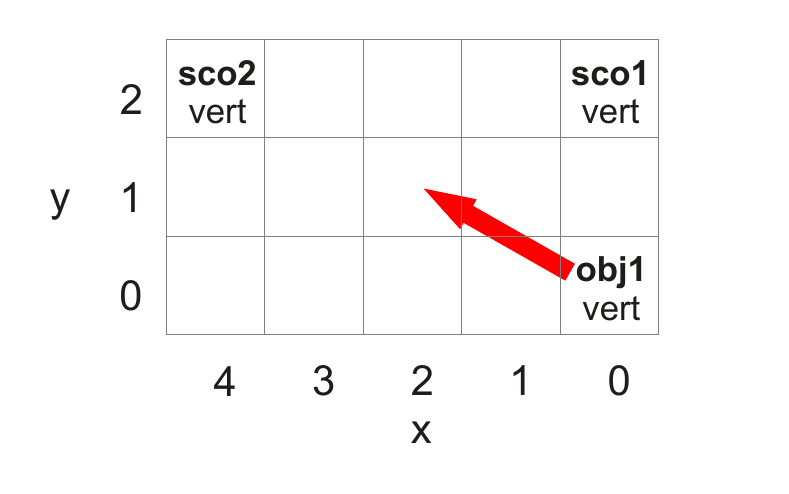} & \includegraphics[scale=0.6]{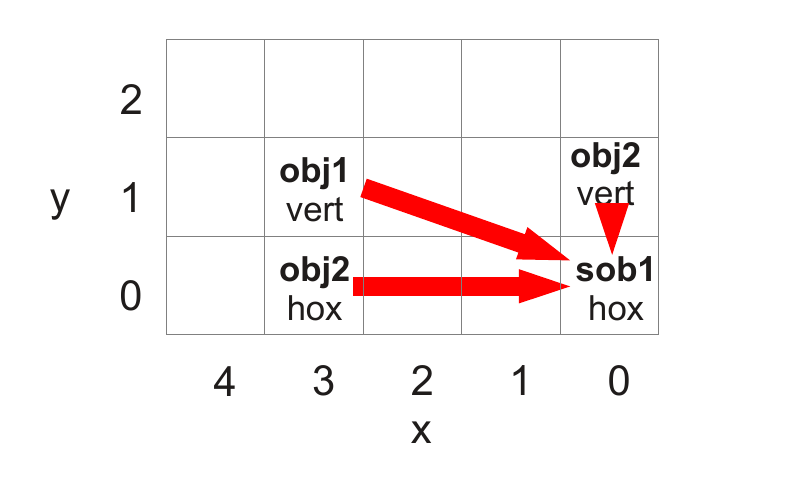}
\end{tabular}
\end{center}\vspace{-0.5cm}
 \caption{Left: Instance 1. Middle: Instance 2. Right: Instance 3. Top: Instance RGB images viewed from the robot-mounted camera before planning. Middle: Depth images from the same camera with overlayed task specification arrows. Bottom: Grid representation of the instance problems as provided to the planner after bottom-up perceptual processing. \textbf{obj1}, \textbf{obj2} and \textbf{obj3}: objects automatically associated with the object from the task specification. \textbf{sco1} and \textbf{sco2}: scene objects extracted during perception but not associated with any task-related object. The initial orientation is one of \textbf{vert}, \textbf{hox}, and \textbf{hoy}. The red arrows represent the task specification.}
\label{all_instances}
 \end{figure}

\paragraph{Instance 2:}
The middle column of Figure \ref{all_instances} shows the second
problem instance. This instance presents a scenario in which the
planning needs to use minimal perceptual information as the objects
in the scene are positioned such that objects apart from the object to be moved
are far enough away that interactions are unlikely. %
The task is to
move the bolt to the center of the workspace (move \verb#obj1# from
\verb#loc_0x0# to \verb#loc_2x1#), which does not require any other
objects to be moved beforehand. For most plans, no stackability or reachability would need to
be checked to determine their feasibility. A sample plan is as follows:

\begin{verbatim}
  0: move(obj1,loc_2x1,horiz_y).
\end{verbatim}

\paragraph{Instance 3:}
The aim of Instance 3 (Figure \ref{all_instances}, rightmost column)
is to present a scenario in which the stackability of objects is a
concern. The task requires all objects to be moved to the front
right of the table, and the planner must infer a plan that brings
the objects there while respecting the stackability constraints that
are inferred using appropriate perceptual information. The aluminium
profiles are safely stacked horizontally and a single bolt is
stackable on top of them as long as it is vertically oriented. The
object currently at the front right of the table need be moved aside
and the remaining objects stacked in the appropriate order.
In general, all object shapes will need to be computed in order to verify a plan. %
 A sample plan would be:

\begin{verbatim}
  0: move(sco1,loc_2x2,horiz_x).
  1: move(obj1,loc_0x0,horiz_y).
  2: move(obj2,loc_0x0,horiz_y).
  3: move(obj3,loc_0x0,vert).
\end{verbatim}

\section{Results}

All experiments were performed on a Linux laptop with a 4-core
2.26GHz Intel i5-430M CPU and 4GB memory. For planning, we use
\ccalc\ (Version 2.0) with the SAT solver mChaff (version spelt3).
For perception, we use an object segmentation and shape recognition
system built on the Kinect RGBD camera and the
Point Cloud Library \cite{rusu_3d_2011} (PCL SVN revision 6849).

\begin{table}[b]%
\vspace{-0.3em}
\scriptsize

\caption{\scriptsize Plan quality for the calculation of the FIRST feasible plan and 100 feasible plans for each of the three problem instances. Reported results are averaged over 5 runs.}%
\centering
\begin{tabular}{l@{\hspace{1cm}}rrrrr@{\hspace{1cm}}rrrr}
\hline\\[-0.4em]
\multicolumn{1}{c}{\multirow{2}{*}{Instance 1}}&  \multicolumn{4}{c}{To FIRST feasible plan} & &\multicolumn{4}{c}{To 100 feasible plans} \\[0.2em]
&  \textsc{None}    &  \textsc{Filt}     & \hspace{1em}\textsc{Pre}  & \hspace{0.7em}\textsc{Repl} & & \textsc{None}  &  \textsc{Filt}     & \hspace{1em}\textsc{Pre}  & \hspace{0.7em}\textsc{Repl} \\
\hline\\
\# perception queries & 0.0 & 2.0 & 3.0 & 1.6 & & 0.0 & 3.0 & 3.0 & 1.4 \\
\# feasible plans & 0.0 & 1.0 & 1.0 & 1.0 & & 7.2 & 100.0 & 100.0 & 100.0 \\
\# infeasible plans & 1.0 & 61.4 & 0.0 & 1.0 & & 92.8 & 505.3 & 0.0 & 1.5 \\[0.4em]
\hline\\[-0.4em]
\multicolumn{1}{c}{\multirow{2}{*}{Instance 2}}&  \multicolumn{4}{c}{To FIRST feasible plan} & &\multicolumn{4}{c}{To 100 feasible  plans} \\[0.2em]
&  \textsc{None}    &  \textsc{Filt}     & \textsc{Pre}  & \textsc{Repl} & & \textsc{None}  &  \textsc{Filt}     & \textsc{Pre}  & \textsc{Repl} \\
\hline\\
\# perception queries  & 0.0 & 0.0 & 3.0 & 0.0 & & 0.0 & 1.2 & 3.0 & 0.8 \\
\# feasible plans & 1.0 & 1.0 & 1.0 & 1.0 & & 94.8 & 100.0 & 100.0 & 100.0 \\
\# infeasible plans & 0.0 & 0.0 & 0.0 & 0.0 & & 5.2 & 6.0 & 0.0 & 0.0 \\[0.4em]
\hline\\[-0.4em]
\multicolumn{1}{c}{\multirow{2}{*}{Instance 3}}&  \multicolumn{4}{c}{To FIRST feasible  plan} & &\multicolumn{4}{c}{To 100 feasible plans} \\[0.2em]
&  \textsc{None}    &  \textsc{Filt}     & \textsc{Pre}  & \textsc{Repl} & & \textsc{None}  &  \textsc{Filt}     & \textsc{Pre}  & \textsc{Repl} \\
\hline\\
\# perception queries  & 0.0 & 4.0 & 4.0 & 4.0 & & 0.0 & 4.0 & 4.0 & 4.0 \\
\# feasible plans & 0.0 & 0.0 & 1.0 & 1.0 & & 0.0 & 0.0 & 100.0 & 100.0 \\
\# infeasible plans & 1.0 & 1759.7 & 0.0 & 1.0 & & 100.0 & 1757.0 & 0.0 & 1.0
\end{tabular}
\label{stats}
\end{table}

\begin{figure}[t]
\vspace{-1.0em}%
\includegraphics[scale=0.45]{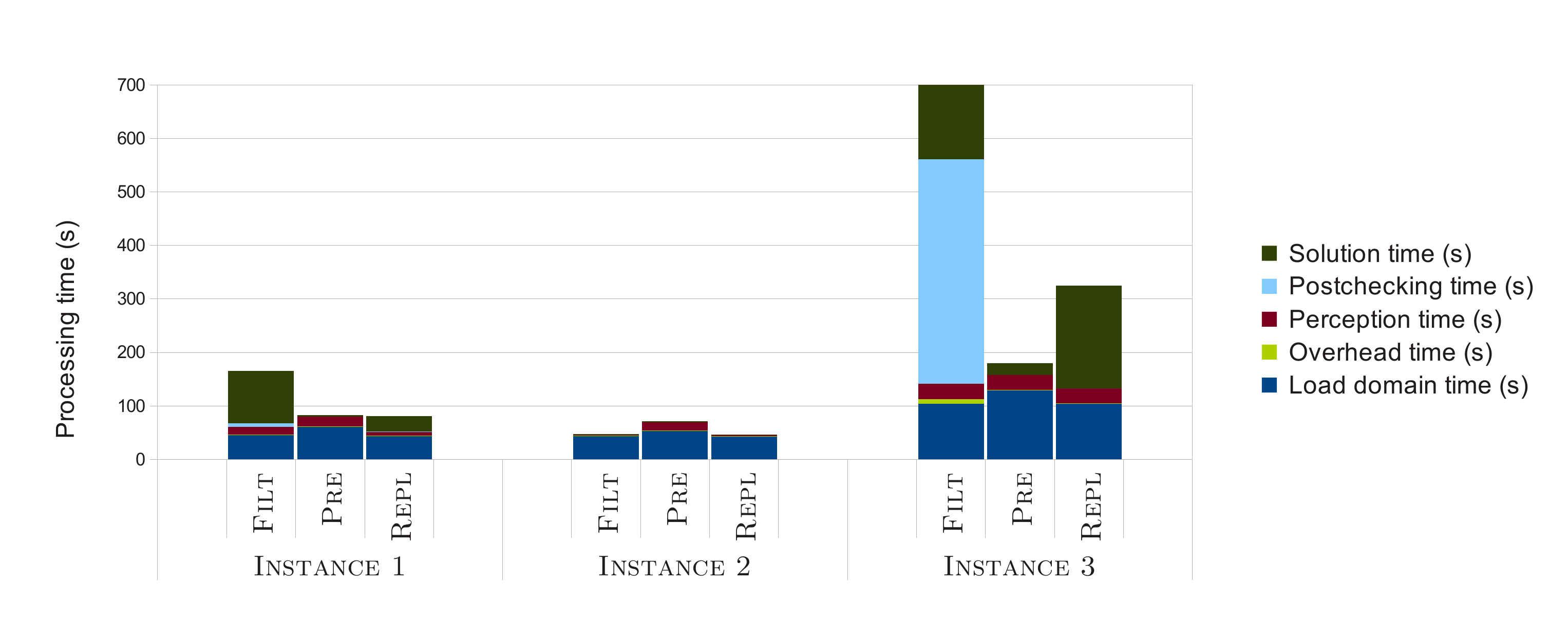}\vspace{-0.5cm}
\caption{\footnotesize Empirical computation time for computing the FIRST feasible plan, averaged over 5 runs.   The column for \textsc{Filt} in instance 3 is cut and it times out after 2000 seconds.}%
\label{empirical1}
\end{figure}

\begin{figure}[t]
\vspace{0.5em}%
\includegraphics[scale=0.45]{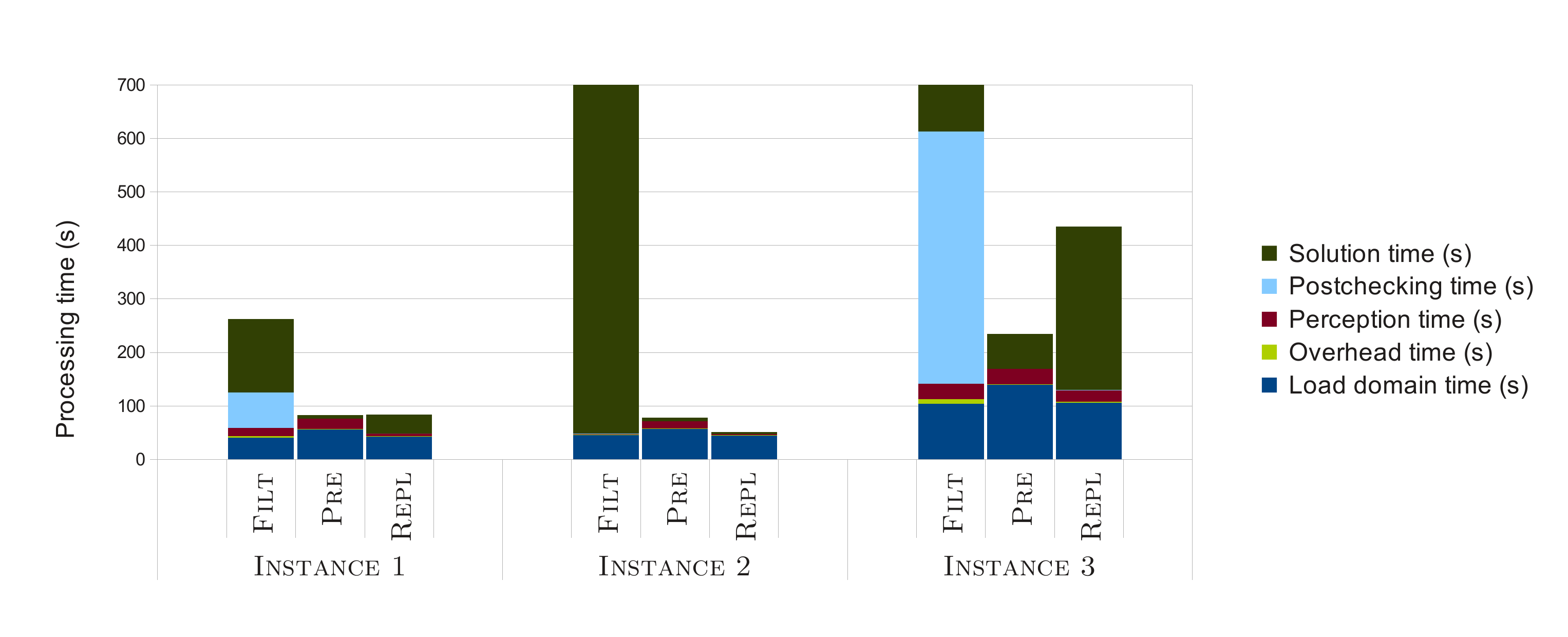}\vspace{-0.5cm}
\caption{\footnotesize Empirical computation time for computing the 100 plans, averaged over 5 runs. The column for \textsc{Filt} in instance 2 and 3 are cut since they time out after 2000 seconds.} %
\label{empirical100}
\end{figure}

For each of the three planning problem instances described above, we ask 1) for the FIRST feasible
plan, and 2) for 100 feasible plans. We analyze the results both from the point of view of plan quality and from the point of view of computation time.
We report the average number of perception queries,
feasible plans and infeasible plans (Table~\ref{stats}), and average computation
times over five runs (Figures~\ref{empirical1} and~\ref{empirical100}). The timeout is set at 2000 seconds.

According to Table~\ref{stats}, considering quality of solutions (the proportion of number of feasible plans with respect to total number of feasible and infeasible plans), we can observe that all integration approaches give better results compared to \textsc{None} in almost all cases. Among the integration approaches, \textsc{Pre} performs the best, since all feasibility checks that are computed in advance are taken into consideration during task planning. On the other hand, \textsc{Filt} performs the worse, since all feasibility checks are done at the very end and no information is conveyed to replanning. \textsc{Repl} performs better than \textsc{Filt} because results of feasibility checks are considered while replanning by means of constraints.

With respect to the amount of perceptual processing necessitated, we can observe that the maximum average number of perception queries takes place in~\textsc{Pre}. Comparing \textsc{Filt} and \textsc{Repl}, more perceptual processing is required by \textsc{Filt} due to a larger variety of infeasible plans checked.

As for computation times, as observed in Figures~\ref{empirical1} and~\ref{empirical100}, due to the large number of infeasible plans generated, \textsc{Filt} takes the maximum time for finding feasible plans. For Instance 3, \textsc{Filt} cannot find a feasible plan within the timeout. Since all feasibility checks can be computed in a relatively short amount of time, \textsc{Pre} performs better than \textsc{Repl} on Instance 3, but performs worse than \textsc{Repl} in Instance~2 where \textsc{Pre} does unnecessary perceptual compuation that does not reduce the time spent calculating plans. On Instance 1, the increased time spent doing perceptual processing and loading the consequently larger domain for \textsc{Pre} is balanced against increased time loading constraints after planning for \textsc{Repl}.

These results are in line with observations by Schüller~et~al.~2013~\cite{schuller_systematic_2013}, where geometric reasoning is integrated with task planning. According to that work, since for some domains computing all feasibility checks in advance is not possible, \textsc{Pre} may not be possible at all. This can happen in domains where all possible perceptual computations take too much time.

\section{Discussion}

We have investigated the usefulness of three different approaches to
integrating planning and perception: \textsc{Pre}  where all
perceptual computations are done before planning to find a feasible
plan, \textsc{Filt} where perceptual computations are done after
planning to check the feasibility of the computed plan, and \textsc{Repl} where
perceptual computations are done after planning to check the
feasibility of the computed plan, and replanning is done with
guidance by incorporating constraints in the planning problem.

Experiments comparing these three approaches
consider three problem instances of a robotic manipulation domain
that involves industrial objects used at RoboCup@Work 2012
competitions.

We have observed that in terms of quality of solutions (that is the rate of infeasible plans produced by the planning module) \textsc{Pre} performs the best. As for computation time, however, \textsc{Pre} also necessitates the maximum number of perceptual queries; the minimum amount of perceptual processing is demanded by \textsc{Repl} which also reduces the upfront computational cost during domain loading but spends extra time due to the overhead of restarting the planner. When a majority of the set of possible feasibility checks are required to find a plan and can be performed quickly in advance, \textsc{Pre} gives the best results; but as the number of external checks needed decreases with respect to the number of possible checks, \textsc{Repl} starts to perform better.
We expect that as the number of objects increase, the number of feasibility checks will increase as well and thus \textsc{Repl} should continue to perform better, as observed in~\cite{schuller_systematic_2013}.

\section*{Acknowledgments}

This work is partially supported by TUBITAK Grant 111E116. Damien J.
Duff is supported by TUBITAK 2216 Research Fellowship.

\bibliographystyle{splncs}
%
% \bibliography{PerceptionIntegration}

%

\label{lastpage}
\end{document}